%% file: main.tex
\documentclass[letterpaper, 10 pt, conference]{ieeeconf}
\IEEEoverridecommandlockouts
\usepackage{amsmath,amssymb,url} % default
\usepackage{balance}

\usepackage[mathscr]{eucal} % adds \mathscr command for Euler script, else overwrites mathcal

\usepackage{hyperref} % gives links for all references

\hypersetup{
	pdftitle={Cooperative Multi-Target Localization With Noisy Sensors},
	pdfauthor={Philip Dames and Vijay Kumar},
	colorlinks=true
}

\usepackage{graphicx} % for pdf, bitmapped graphics files
\graphicspath{{./figures/}}
\usepackage[caption=false, font=footnotesize]{subfig} % options for ieeeconf

\newcommand{\ie}{\textit{i.e.,}\ }
\newcommand{\eg}{\textit{e.g.,}\ } % the \ ensures latex does not treat command as end of sentence
\newcommand{\pbar}{\overline{p}}

\newcommand{\E}[1]{\mathbb{E}[#1]} % expectation

\newcommand{\R}{\mathbb{R}} % real numbers

\renewcommand{\L}{\mathcal{L}}
\newcommand{\G}{\mathcal{G}}
\newcommand{\X}{\mathscr{X}}
\newcommand{\Z}{\mathcal{Z}}

\DeclareMathOperator*{\argmax}{arg\,max}

\title{\LARGE \bf Technical Report: Cooperative Multi-Target \\ Localization With Noisy Sensors}
\author{Philip Dames, Vijay Kumar % <-this % stops a space
\thanks{This work was funded in part by ONR Grant N00014-07-1-0829 and AFOSR Grant FA9550-10-1-0567.  
Philip Dames was supported by the Department of Defense (DoD) through the National Defense Science \& Engineering Graduate Fellowship (NDSEG) Program.}% <-this % stops a space
\thanks{P. Dames and V. Kumar are with The GRASP Lab,
    University of Pennsylvania, Philadelphia, PA 19104, USA 
    {\tt\small\{pdames,\allowbreak kumar\}\allowbreak @seas.\allowbreak upenn.edu}}%
}

\date{\today}

\IEEEaftertitletext{\vspace*{-3.5mm}} % adjust spacing after title

\begin{document}
\maketitle
%%%%%%%%%%%%%%%%%%%%%%%%%%%%%%%%%%%%%%%%%%%%%%%%%%%%%%%%%%%

%%%%%%%%%%%%%%%%%%%%%%%%%%%%%%%%%%%%%%%%%%%%%%%%%%%%%%%%%%%
\begin{abstract}
This technical report is an extended version of the paper `Cooperative Multi-Target Localization With Noisy Sensors' accepted to the 2013 IEEE International Conference on Robotics and Automation (ICRA).

This paper addresses the task of searching for an unknown number of static targets within a known obstacle map using a team of mobile robots equipped with noisy, limited field-of-view sensors.
Such sensors may fail to detect a subset of the visible targets or return false positive detections.
These measurement sets are used to localize the targets using the Probability Hypothesis Density, or PHD, filter.
Robots communicate with each other on a local peer-to-peer basis and with a server or the cloud via access points, exchanging measurements and poses to update their belief about the targets and plan future actions.
The server provides a mechanism to collect and synthesize information from all robots and to share the global, albeit time-delayed, belief state to robots near access points.
We design a decentralized control scheme that exploits this communication architecture and the PHD representation of the belief state.
Specifically, robots move to maximize mutual information between the target set and measurements, both self-collected and those available by accessing the server, balancing local exploration with sharing knowledge across the team.
Furthermore, robots coordinate their actions with other robots exploring the same local region of the environment.
\end{abstract}
%%%%%%%%%%%%%%%%%%%%%%%%%%%%%%%%%%%%%%%%%%%%%%%%%%%%%%%%%%%

%%%%%%%%%%%%%%%%%%%%%%%%%%%%%%%%%%%%%%%%%%%%%%%%%%%%%%%%%%%
\section{Introduction}
\label{sec:introduction}
%%%%%%%%%%%%%%%%%%%%%%%%%%%%%%%%%%%%%%%%%%%%%%%%%%%%%%%%%%%
Teams of mobile robots are often used to gather information; to detect, localize and track targets; and to map the environment.
The presence of multiple robots allows for simultaneous exploration of disjoint areas of the environment and cooperative viewing of the same location from multiple vantage points, but raises several key questions not present in single-robot scenarios.
Namely how should robots communicate with each other and how should robots coordinate their actions?
This paper seeks to answer these questions by drawing upon work in robot network architecture, information-based control, and multi-target localization.

In particular we examine the problem of searching for an unknown number of static targets within a known map.
Examples of situations where such a task is applicable include surveillance, security, and monitoring, all of which take place in locations where there may be an existing communication infrastructure, \eg a wireless network or intermittent satellite communication, that the team can leverage.

%%%%%%%%%%%%%%%%%%%%%%%%%%%%%%
\begin{figure}[t]
\centering
\includegraphics[width=0.9\columnwidth]{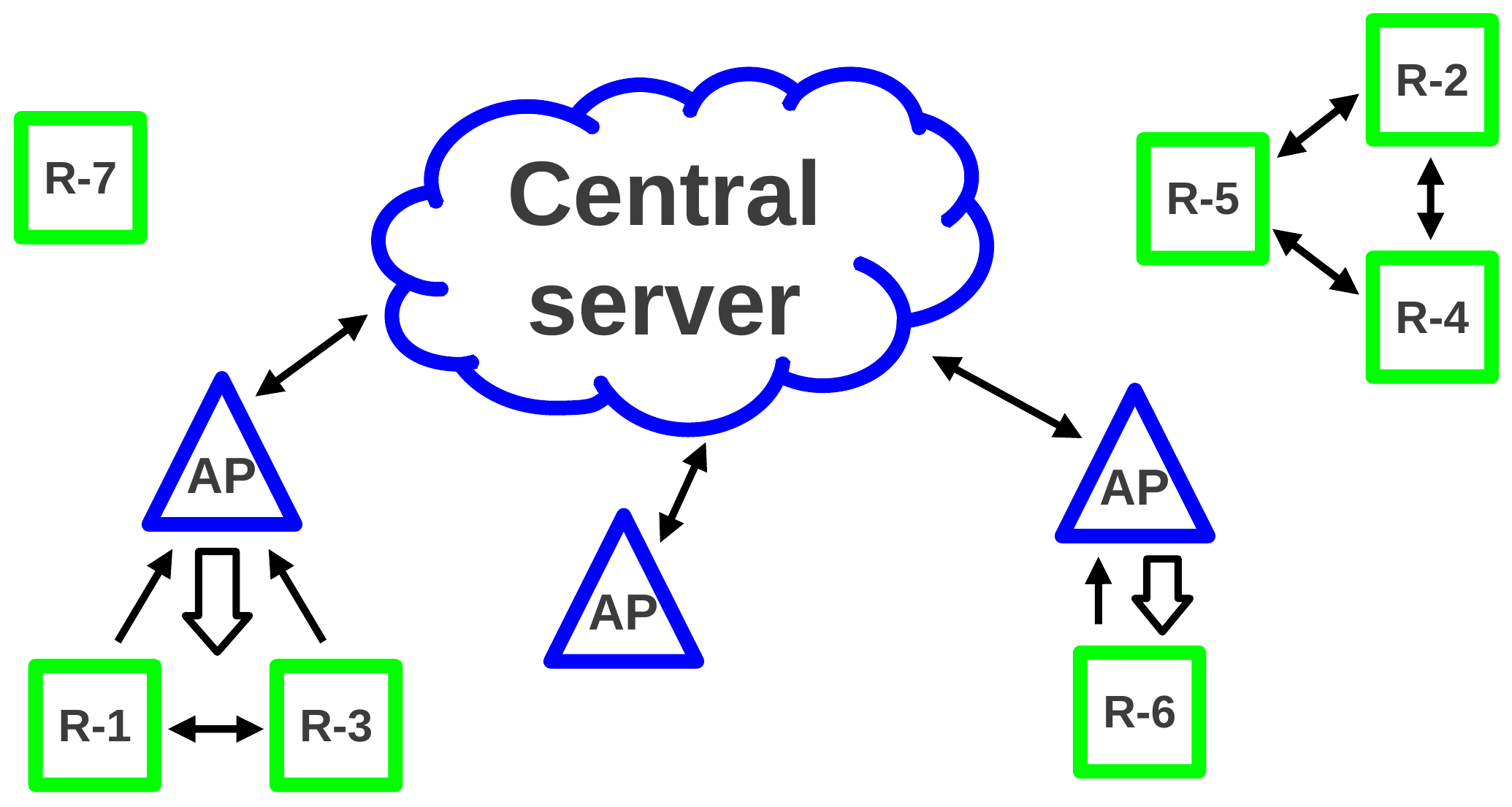} \vspace*{-2mm}
\caption{Diagram of the network structure.
Robots (green squares) are able to communicate on a peer-to-peer basis with nearby robots as well as access the central server through access points (blue triangles).
The communication links originating from robots are all relatively low-bandwidth while the downlink from the server may be higher bandwidth.
}
\label{fig:networkstructure}
\vspace*{-5mm}
\end{figure}
%%%%%%%%%%%%%%%%%%%%%%%%%%%%%%

The need for a communication architecture is central to the performance of a cooperative robotic team, yet must take into account the limited capabilities (\eg communication range and bandwidth) of each robot while allowing robots to exchange information in a consistent way.
A centralized approach will not work over large scale environments where not all robots will be able to communicate with one another.
One common decentralized architecture is Decentralized Data Fusion (DDF), first described by Grime and Durrant-Whyte \cite{grime1994data}, in which each robot manages its own copy of the joint belief and aggregates data from the other robots through channel filters which only admit information that is distinct from their current belief.
The DDF framework is particularly amenable to Gaussian beliefs as the information form of the Kalman filter allows for efficient, low-bandwidth updates.
However, more complicated belief representations often require overly conservative approaches to data fusion.

Our solution takes the best of each of these approaches, allowing robots to communicate on a peer-to-peer basis in a decentralized fashion while also including communication with a centralized server or a cloud which robots may access via the existing network infrastructure in the environment.
This idea of robots relying on information from a server has been called cloud robotics and has recently generated quite a bit of excitement \cite{guizzo2011cloud,guizzo2011robots}.
A similar idea was also used for estimation and control of groups of robots by Michael, Fink, and Kumar \cite{michael2011architecture} where an Asymmetric Broadcast Control (ABC) was used to synthesize locally derived information and provide low-resolution global information to the group.
The asymmetry is in the communication between the robots and the server.
Uploads from robots are low-bandwidth by nature but downloads involving global information may require higher bandwidth.
Robots are not required to constantly communicate with the central server or cloud, instead they opportunistically upload and retrieve information based on their physical proximity to access points.
This is shown in Fig.~\ref{fig:networkstructure} where robots may have one or more communication links and can trade off the benefits of accessing the server compared to taking further local measurements.

One common approach to robot control for active estimation is to maximize mutual information between the target locations and the robots' measurements.
Both Grocholsky \cite{grocholsky2002thesis} and Cole \cite{cole2009thesis} consider information-theoretic control of robot teams for exploration and tracking tasks using the DDF architecture to handle inter-agent communication.
In particular, Cole \cite{cole2009thesis} examines the scenario where the number of targets is unknown, deriving equations similar to those of the PHD filter but using a very conservative data fusion approach.
Stranders, et al \cite{stranders2009decentralised} and Delle Fave, et al \cite{delle2012deploying} use the max-sum algorithm for decentralized control computations and DDF to share beliefs about target locations.

When robots have noisy limited field-of-view sensors, it is often necessary to use target models with non-parametric distributions and to consider the possibility of false positive detections and errors in data association.
In our previous work, we have developed control policies that use gradients of mutual information to drive mobile robots with binary sensors to search for targets without making assumptions on data association or the underlying distribution \cite{SchwagerISRR11Hazards,DamesCDC12Decentralized}.
This approach is based upon finite set statistics, the probabilistic framework used to derive the Probability Hypothesis Density, or PHD, filter.
A brief overview of finite set statistics is provided in Sec.~\ref{sec:estimation}.
However, most of this work considers static sensors, with the only other papers dealing with control for estimation by Ristic, et al \cite{ristic2010sensor,ristic2011note}, who use R\'{e}nyi divergence, a generalization to mutual information, to drive a single robot to search for targets.
The R\'{e}nyi divergence is computed using Monte Carlo integration, while this paper utilizes analytic approximations to mutual information.

This paper presents a decentralized control architecture founded upon the ideas of information gathering, synthesis, and dissemination.
Gathering is done using a team of mobile sensors, the only strong assumption being that robots are able to localize themselves and navigate without noise.
The data is then incorporated into the robot's belief through the PHD filter, making no additional assumptions on the targets' spatial or cardinality distributions.
The synthesis of peer-to-peer and cloud information is done in a principled way, synchronizing the beliefs of robots and ensuring no data is double counted as it is exchanged.
Mutual information is used to balance the benefits of obtaining information by direct observation of the environment or by downloading from the server, merging the objectives of gathering and disseminating information into a single control law.

%%%%%%%%%%%%%%%%%%%%%%%%%%%%%%%%%%%%%%%%%%%%%%%%%%%%%%%%%%%
\section{Modelling}
\label{sec:problemformulation}
%%%%%%%%%%%%%%%%%%%%%%%%%%%%%%%%%%%%%%%%%%%%%%%%%%%%%%%%%%%
In this work a team of $N$ autonomous robots explore a closed environment $E \subset \R^2$.
A list of symbols is given in Table~\ref{tab:symbols}.
The robots seek to localize a set of $J$ stationary targets $X$, where both the cardinality ($|X| = J$) and locations $x$ of the targets are unknown a priori.
The notation $|\cdot|$ indicates the size of a set.

%%%%%%%%%%%%%%%%%%%%%%%%%%%%%%
\begin{table}[tH]
%\vspace*{2mm}
\centering
\begin{tabular}{c l}
$q_i \in E$ & Position of robot $i$ in environment \\
$x \in X$ & Target location in target set \\
$p_d(x; q)$ & Probability of a robot at $q$ detecting a target at $x$ \\
$F_i$ & Sensor footprint of robot $i$ \\
$z \in Z$ & Measurement in measurement set \\
$g(z \mid x; q)$ & Single-target measurement model \\
$\kappa(z), \, \mu$ & Clutter PHD, expected cardinality \\
$D(x), \, \lambda$ & Target PHD, expected cardinality \\
$C$ & Coalition of robots \\
%$\alpha(C)$ & Expected cardinality of non-detected targets by $C$
\end{tabular}
\caption{Table of symbols}
\label{tab:symbols}
\vspace*{-8mm}
\end{table}
%%%%%%%%%%%%%%%%%%%%%%%%%%%%%%

%%%%%%%%%%%%%%%%%%%%%%%%%%%%%%
\subsection{Sensing}
\label{sec:sensing}
%%%%%%%%%%%%%%%%%%%%%%%%%%%%%%
Each robot is equipped with a noisy sensor that returns a set of measurements $Z^t$ at each time step $t$, which is then used to update the estimate of the target set.
There is also the possibility that some targets are missed due to sensor failure (\ie false negatives) and that measurements may be due to clutter within the environment (\ie false positives).

Use of the PHD filter (see Sec.~\ref{sec:estimation}) requires probabilistic models of the probability of detecting true targets, a single-target measurement model, and the clutter detection probabilities.
The probability of a robot at $q$ detecting a target at $x$, $p_d(x; q)$, depends upon the robot position though we will often omit the dependence of $q$ for notational compactness.
Robot $i$ may only detect targets within its sensor footprint $F_i$, \ie $p_d(x) = 0 \quad \forall x \notin F_i$.

If a target at $x$ is detected within the footprint, then a measurement is returned according to the model $g(z \mid x, q)$, though again the dependence on $q$ is omitted.
Note that in this work the term measurement refers to a high-level reading rather than the raw sensor data, \eg the output of a target classifier over an image instead of the image itself.

Finally, we must take into account the possibility of returning false positive measurements.
In particular, we assume that the clutter detections are well modelled by a Poisson random finite set so we need only the PHD $\kappa(z)$, where $\mu \triangleq \int \kappa(z) \, dz$ is the expected cardinality.
In the absence of a priori information about likely clutter locations let $\kappa(z)$ be piecewise constant such that $\kappa(z) = 0$ for all $z$ that could not have originated from a target within the sensor footprint.

%%%%%%%%%%%%%%%%%%%%%%%%%%%%%%
\subsection{Communication}
\label{sec:communication}
%%%%%%%%%%%%%%%%%%%%%%%%%%%%%%
As robots explore the environment, they store a local history of messages, where messages consist of (position, measurement set) pairs.
This history will be shared with other robots directly over peer-to-peer links, and indirectly through the central server, to aid in exploration.
The central server has $A$ stationary access points located in the environment at $s_1, \ldots s_A$, at which robots upload messages and download the latest PHD from the server, $D_s(\cdot)$.

Robot-server communication, as previously noted~\cite{michael2011architecture}, is asymmetric in the bandwidth.
When a robot is within communication range of an access point, the robot uploads its message history since the last check-in, waits while the server uses these messages to update its PHD $D_s$, and receives the resulting PHD from the server.
This PHD $D_s$ replaces the robot's PHD as it includes all of the robot's own message history as well as all information uploaded by other robots prior to the current time.

On the other hand, robot-robot communication is symmetric.
Here robots form coalitions, which are connected components of a communication graph with edges between robots that are able to communicate.
Robots then simply exchange their most recent messages with all other robots in the coalition.
These messages are then used to update the PHD.
This framework allows robots to jointly explore the environment while not double-counting any information, as communication with the central server overwrites the peer-to-peer updates.

%%%%%%%%%%%%%%%%%%%%%%%%%%%%%%%%%%%%%%%%%%%%%%%%%%%%%%%%%%%
\section{Estimation}
\label{sec:estimation}
%%%%%%%%%%%%%%%%%%%%%%%%%%%%%%%%%%%%%%%%%%%%%%%%%%%%%%%%%%%
In this work we will use an estimation method based on finite set statistics, a probabilistic framework that deals with uncertainty in both the cardinality and positions of targets in a principled fashion.

%%%%%%%%%%%%%%%%%%%%%%%%%%%%%%
\subsection{Background}
\label{sec:fisst}
%%%%%%%%%%%%%%%%%%%%%%%%%%%%%%
Finite set statistics (FISST) was first applied to engineering problems by Mahler \cite{mahler2003multitarget} where he considered radar-based tracking of an unknown number of mobile target and has recently been adopted in the robotics community for feature-based mapping and SLAM by Mullane, et al \cite{mullane2011random,mullane2011randompaper}.
The key distinction between FISST and traditional estimation methodologies is that FISST is based on the concept of a \emph{random finite set} (RFS), a set containing a randomly varying number of random vectors.
In the context of target tracking, a realization of the RFS gives the number (\ie set cardinality) and position (\ie vectors in the set) of the targets.

It is also important to note that sets do not provide any label or ordering to the targets, as sets are equivalent under permutation of the elements.
This allows FISST to avoid one issue that arises in multitarget tracking, that of data association, \ie matching measurements to individual targets, by averaging over all possible data associations.

One issue that arises is that there is no notion of addition with sets, so care must be taken when performing integration over a RFS.
To this end, Mahler defines the set integral
\begin{equation}
	\label{eq:setintegral}
	\int f(X) \, \delta X \triangleq \sum_{n=0}^\infty \frac{1}{n!} \int f(\{x_1, \ldots, x_n\}) \, dx_1 \ldots dx_n
\end{equation}
where $f(X) = f(\pi(X))$ for any permutation $\pi$ of a set $X$.

The PHD is the first statistical moment of the distribution over RFSs.
It takes the form of a target density function over the environment with the property that the integral over any region gives the expected number of targets in that region.
Note that this is \emph{not} a probability density function.

%%%%%%%%%%%%%%%%%%%%%%%%%%%%%%
\subsection{PHD Filter}
\label{sec:phdfilter}
%%%%%%%%%%%%%%%%%%%%%%%%%%%%%%
The PHD filter is a set of computationally tractable recursive equations to update the probability hypothesis density, which is the first statistical moment of a distribution over RFSs.
In general the PHD filter makes the following assumptions about the targets and robots:
\begin{itemize}
\item targets move and generate measurements independently
\item new and surviving targets are independent
\item the clutter RFS is Poisson and independent of measurements generated by true targets
\item the predicted target RFS is Poisson.
\end{itemize}
Here the term clutter is synonymous with false positive detections.
Under these assumptions, the optimal Poisson approximation (a RFS is said to be Poisson if the number of targets is Poisson and target locations are i.i.d.) of the multitarget density is
\begin{equation}
	\label{eq:poissonrfs}
	p(X) = e^{-\lambda} \prod_{x \in X} D(x)
\end{equation}
where $D(\cdot)$ is the PHD and $\lambda \triangleq \int D(x) \, dx$ is the parameter of the Poisson distribution.
This comes from Theorem 4 by Mahler \cite{mahler2003multitarget}, who goes on to derive the PHD prediction step 
\begin{equation}
	\label{eq:phdpredict}
	D^{t \mid t-1}(x) = \gamma(x) + \int p_s(\xi) f(x \mid \xi) D^{t-1 \mid t-1}(\xi) \, d\xi
\end{equation}
where $p_s(x)$ is the probability of a target surviving between time steps, $f(x \mid \xi)$ is the target motion model, and $\gamma(x)$ is the PHD of new targets entering the environment, and the PHD update step:
\begin{align}
	D^{t \mid t}(x) =& \, \L(x, Z^t, D^{t \mid t-1}) \, D^{t \mid t-1}(x) \label{eq:phdupdate} \\
	\L(x, Z, D) =& \, \overline{p}_d(x) + \sum_{z \in Z} \frac{p_d(x) g(z \mid x)}{\kappa(z) + \int p_d(\xi) g(z \mid \xi) D(\xi) \, d\xi}. \label{eq:updateoperator}
\end{align}
The $\overline{\cdot}$ notation will be used throughout to indicate the additive complement of a probability (\eg $\overline{p}_d(x) = 1 - p_d(x)$) and superscript $t$ to represent the time index.

%%%%%%%%%%%%%%%%%%%%%%%%%%%%%%
\subsection{Further Assumptions}
\label{sec:assumptions}
%%%%%%%%%%%%%%%%%%%%%%%%%%%%%%
We focus on the case of stationary targets so that $p_s = 1$, $f(x \mid \xi) = \delta(x,\xi)$ is the identity map (where $\delta(x,\xi)$ is the Dirac delta), and $\gamma = 0$.
Note that \eqref{eq:phdpredict} then simplifies to $D^{t \mid t-1}(\cdot) = D^{t-1 \mid t-1}(\cdot)$ so we adopt the shorthand notation $D^t(\cdot) \triangleq D^{t \mid t}(\cdot)$ and only~\eqref{eq:phdupdate} is needed to update the belief.

The PHD is often represented as a mixture of Gaussians or as a weighted particle set, as was done by Vo and Ma \cite{vo2006gaussian} and Vo, Singh, and Doucet \cite{vo2003sequential}, respectively.
We elect to take the latter approach, representing the PHD as a set of stationary weighted particles.
In other words, the PHD is
\begin{equation}
D(x) \approx \sum_{p=1}^{P} w_p \delta(x,x_p)
\end{equation}
where $w_p$ is the weight of the particle at position $x_p$, $x_p$ does not depend on time, and $P$ is the number of particles.
There may also be an asymmetry in the robot and server PHDs, with the robots having a coarser resolution due to limited computational resources.

This framework was extended to consider arbitrary distributions over target number by Mahler \cite{mahler2007cphd} and all results in this work may be easily extended to this model.
We use the standard PHD filter, as novel estimation methods are not the focus of the work.

%%%%%%%%%%%%%%%%%%%%%%%%%%%%%%%%%%%%%%%%%%%%%%%%%%%%%%%%%%%
\section{Control}
\label{sec:control}
%%%%%%%%%%%%%%%%%%%%%%%%%%%%%%%%%%%%%%%%%%%%%%%%%%%%%%%%%%%
In order to quickly localize the unknown target set, robots move in such a way as to maximize mutual information between sensor readings and the target set.
Mutual information is an information theoretic quantity which quantifies the amount of information that may be gained about one random variable (\eg target locations) by observing another (\eg measurements).

\begin{figure}[ht]
\vspace*{-5mm}
\centering
\hfill
\subfloat[Example control graph $\G$ for an omnidirectional robot in a small environment.
Nodes are denoted with circles and the shaded nodes are neighbors of the boxed nodes.]{
	\includegraphics[width=0.54\columnwidth]{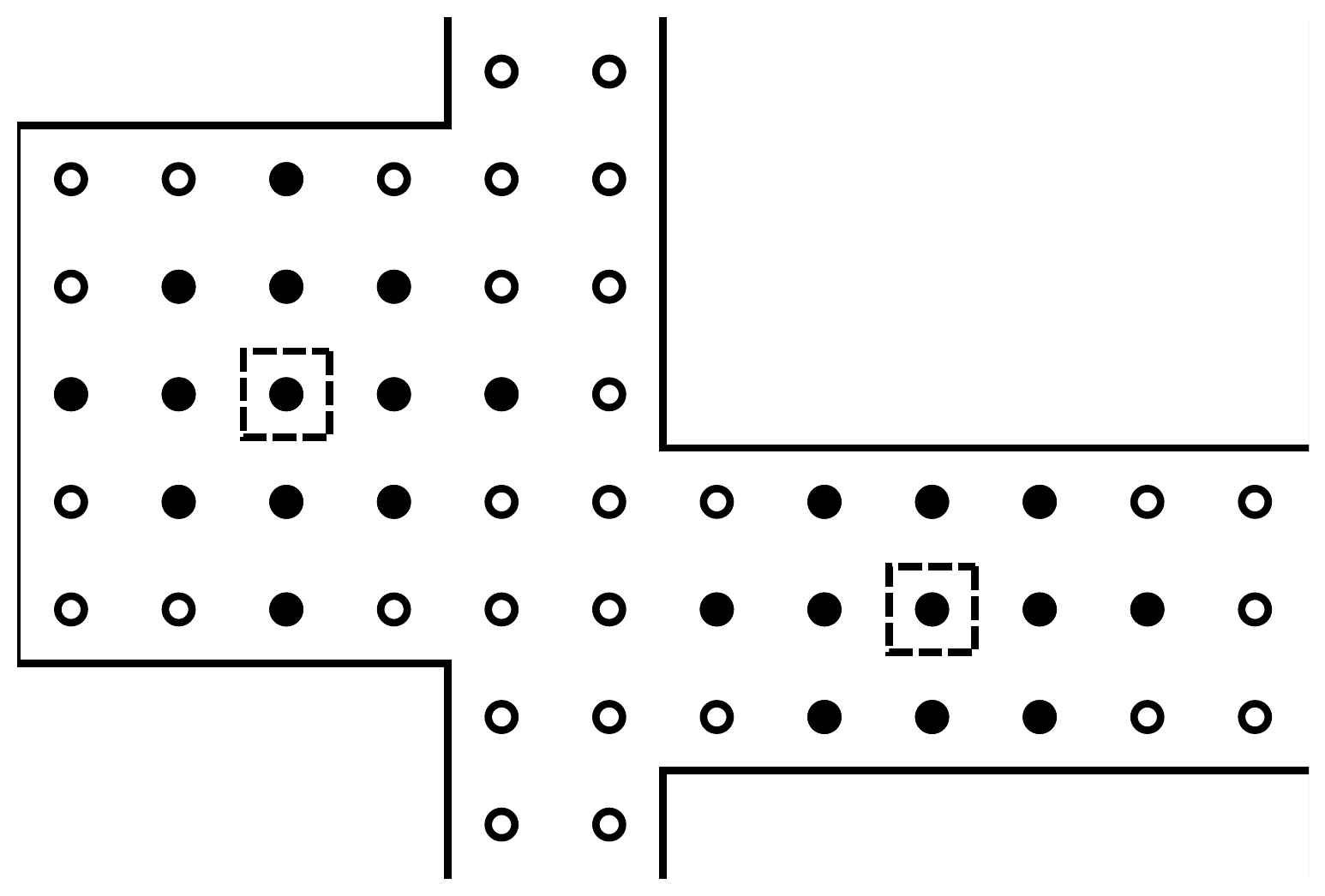}
	\label{fig:examplegraph}}
\hfill \hfill
\subfloat[Finite state machine of the three control modes.]{
	\includegraphics[width=0.33\columnwidth]{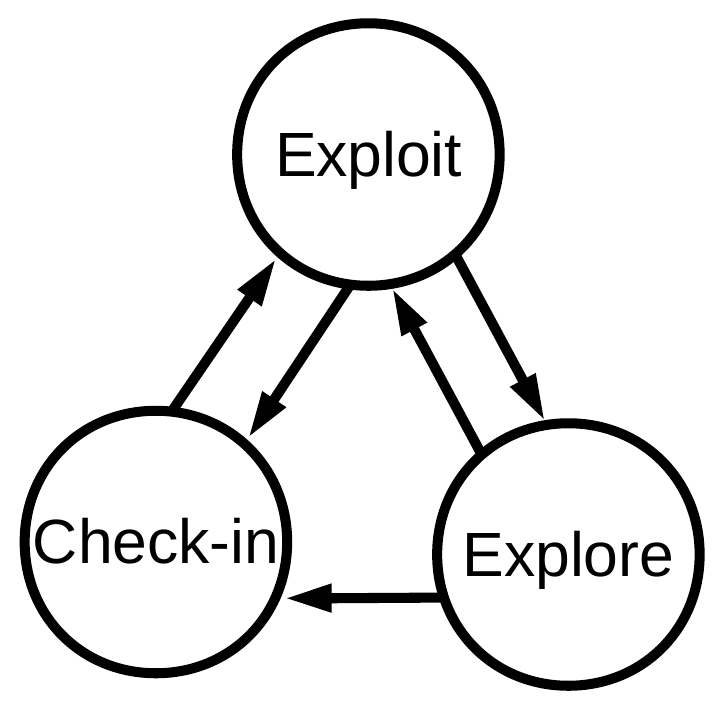}
	\label{fig:finitestatemachine}}
\hfill
\caption{Visualizations of control graph $\G$ and control mode transitions.}
\vspace*{-2mm}
\end{figure}

In contrast to work in POMDPs and SLAM, we assume that robots are able to accurately localize themselves within the environment and have no uncertainty in the execution of actions.
We also restrict the motion of robots to a discrete set of points within the environment, described by the nodes of a graph $\G$.
Edges connect nodes that are reachable within a single time step from the current location, based upon the kinematic restrictions of the robots, and the construction of the graph depends upon the type of sensor being used.
For example, robots equipped with a directional camera will have to search over a larger set of actions than robots with radio range sensors because cameras have an orientation to them.
An example of such a graph is shown in Fig.~\ref{fig:examplegraph}.
Since the map, and thus the graph, are known a priori, it is advantageous to use the Floyd-Warshall algorithm to pre-compute the all-pairs shortest paths between nodes on the graph to allow for fast online look-up of distances and paths.

%%%%%%%%%%%%%%%%%%%%%%%%%%%%%%
%%%%%%%%%%%%%%%%%%%%%%%%%%%%%%

%%%%%%%%%%%%%%%%%%%%%%%%%%%%%%
\subsection{Binary Sensor Model}
\label{sec:binarymodel}
%%%%%%%%%%%%%%%%%%%%%%%%%%%%%%
Due to prohibitively expensive control computations, we must use a simplified sensor model as compared to that used in the PHD updates.
In particular we consider a binary sensor modality which returns $0$ when the measurement set is empty and $1$ otherwise.
The intuition behind this choice is that the robots will move to locations which have a high likelihood of detecting targets, thus gaining information.

The derivation of the sensor model is straightforward.
Note that the only way to have an empty measurement set is to not see all targets and to not have any false positives, so
\begin{equation}
p_b(z=0 \mid X) = p_K(0) \prod_{x \in X} \overline{p}_d(x)
	= e^{-\mu} \prod_{x \in X} \overline{p}_d(x)
\end{equation}
where $p_K(n) = e^{-\mu} \mu^n/n!$ is the probability of $n$ clutter readings and the subscript $b$ denotes the use of the binary model.
Then $p_b(z = 1 \mid X) = \overline{p}_b(z = 0 \mid X)$.

%%%%%%%%%%%%%%%%%%%%%%%%%%%%%%
\subsection{Control Law}
\label{sec:controllaw}
%%%%%%%%%%%%%%%%%%%%%%%%%%%%%%
There are three possible motion modalities for the sensors, the choice of which depends upon the recent history of the robot actions: Explore, Check-in, and Exploit.
A finite state machine, Fig.~\ref{fig:finitestatemachine}, shows the possible mode transitions.
For both the Explore and Check-in modes, robots select a goal node $g \in \G$ and look up the pre-computed shortest path to $g$ from the current location $q_i^t$.
In general these paths require many individual motions due to the limitations on speed, so that robots collect measurements along the way but do not react on them.

\paragraph*{Explore}
In this mode, robots seek out promising areas to search for targets.
This can be done in many ways, both deterministically (\eg lawnmower pattern or maximal coverage) or stochastically.
We opt for the latter, driving robots to a random location within the environment if they have become ``stuck'', \ie when it has not left a small neighborhood $U \subset E$ for a certain number of time steps $T_S$.
This typically happens if a robot has spent many time steps exploring the same region so that there is little uncertainty in the local belief.

\paragraph*{Check-in}
In order to keep the belief in the server (which may be monitored by a human operator) up-to-date and the robots' beliefs somewhat synchronized, robots are required to check-in with the server at least every $T_C$ time steps.
This behavior may be removed by setting $T_C = \infty$.

\paragraph*{Exploit}
If a robot enters the Exploit mode, it will look for nearby robots so that they may coordinate their actions and explore more quickly.
To this end, we redefine a coalition to be a connected component of the control graph, where edges indicate that robots can communicate \emph{and} their sensor footprints overlap, \ie $F_i \cap F_j \neq \varnothing \Rightarrow i,j \in C$.
Each coalition then elects as its leader the robot that has most recently checked in with the server as the leader.
The leader then plans the joint action of all robots in that coalition using its own PHD, which in general differs from that of other robot's, in order to reduce redundancies robot motions.

To plan such an action, the robot maximizes the sum of two components of mutual information in~\eqref{eq:controllaw}.
One due to the measurements taken in the local region of the environment by members of the coalition and the other for measurements uploaded to the server since the last check-in time.
Let $Q_C^t$ be the current joint configuration of robots in a generic coalition $C$, then $Q_C^{t+1} \in \G_C^{t+1} = \prod_{j \in C} \G_j^{t+1}$, where $\G_i^{t+1} \supseteq N(q_i^t) \ni q_i^t$ and $N(q_i^t)$ are the neighbors of $q_i^t$ in $\G$.
The control law is given by
\begin{equation}
	\label{eq:controllaw}
	Q_{C_i}^{t+1} = \argmax_{Q \in \G_{C_i}^{t+1}} \; I[\X, \Z_{C_i}; Q] + I[\X, \Z_s; q_i]
\end{equation}
where $Z_{C_i}$ is the set of binary measurements for coalition $C_i$, $Z_s$ is the set of measurements available at the server, and the semicolon denotes the dependence of information upon the robots' positions.
Robot $i$ then moves to $q_i^{t+1} \in Q_{C_i}^{t+1}$.
We have derived closed form expressions for the mutual information shown in~\eqref{eq:controllaw}, though for brevity only the results are shown here.
The derivation for a single robot may be found in Appendix~\ref{app:MIsingle} and for multiple robots in Appendix~\ref{app:MImultiple}.

The mutual information due to the local measurements is:
\begin{equation}
	\label{eq:coalitioninfo}
	I[\X, \Z_{C_i}; Q] = H[\Z_{C_i}] - H[\Z_{C_i} \mid \X]
\end{equation}
\begin{equation}
	\label{eq:entropy}
	H[\Z_{C_i}] = -\sum_{Z \in \{0, 1\}^{|C_i|}} p_b(Z) \log p_b(Z)
\end{equation}
\begin{equation}
	\label{eq:coalitioncondent}
	H[\Z_{C_i} \mid \X] = \sum_{j \in C} H[\Z_j \mid \X]
\end{equation}
\begin{equation}
	\label{eq:singlecondent}
	H[\Z_j \mid \X] = -\sum_{z_j \in \{0,1\}} \int p_b(z_j, X) \log p_b(z_j \mid X) \, \delta X.
\end{equation}
Note that~\eqref{eq:coalitioncondent} comes from the conditional independence of measurements between robots while the remainder are standard definitions of mutual information ($I$) and entropy ($H$).
The expression for $p(Z)$ is
\begin{equation}
	\label{eq:measurementlikelihood}
	p(Z) = \, \sum_{C' \subseteq C^1} (-1)^{|C'|} e^{-(\lambda - \alpha(C^0 \cup C') + \mu |C^0 \cup C'|)}
\end{equation}
where $C^z = \{j \in C \mid z_j = z\}$ (\ie $C^z$ is the subset of robots in coalition $C$ with measurement $z$) and 
\begin{equation}
	\alpha(C) = \int \prod_{j \in C} \overline{p}_d(x; q_j) D(x) \, dx.
\end{equation}
Finally, the conditional entropy is given by
\begin{equation}
	\label{eq:conditionalentropy}
	H[\Z_j \mid \X] = e^{-(\lambda - \alpha(j) + \mu)} (\mu + \beta)
		- \sum_{\ell	 = 1}^\infty c_\ell e^{-(\lambda - \alpha(j^\ell) + \ell \mu)}
\end{equation}
where $c_1 = -1$, $c_\ell = 1/(\ell (\ell - 1))$, $j^\ell$ is the set containing $\ell$ copies of $j$, and:
\begin{equation}
	\beta = - \int \overline{p}_d(x) D(x) \log \overline{p}_d(x) \, dx.
\end{equation}
The summation and coefficients $c_\ell$ in~\eqref{eq:conditionalentropy} are due to taking a Taylor series of $\log$ about 0.
Note that truncating this to a finite sum means that~\eqref{eq:coalitioninfo} is no longer bounded from below by zero (a standard property of mutual information).
However, we have noticed through empirical simulations that the number of terms used beyond $\ell = 20$ has minimal effect on the control decision made, so we truncate there to minimize computational cost.

The mutual information due to possible measurements in the server is more difficult to model, as the number of such measurements and the locations at which they were taken are unknown until the robot has reached an access point.
For this reason we assume $p_d$ to be independent of the target position.
Let the nominal value be the fraction of the environment covered by the other $N-1$ sensors, which when $|F| \ll |E|$ we have $(N-1)|F|/|E|$.
This is then discounted by some monotonically decreasing function, $h$, of the distance from the robot from the nearest access point such that $h(0) = 1$, $h(\infty) \geq 0$, and
\begin{equation}
	p_d(q) = (N-1) \frac{|F|}{|E|} h\left(\min_{a \in \{ 1, \ldots A \} } d_{\G}(q, g_a) \right)
\end{equation}
where $g_a \in \G$ is the node closest to the access point at $s_a$ and $d_{\G}(q, g)$ is the distance along the graph between $q, g$.

Since robots do not know the locations at which measurements were taken, we assume that measurements are independent of one another (which is true provided that sensor footprints do not overlap).
In this case, mutual information may be written as
\begin{equation}
I[\X, \Z_s; q_i] = \E{m} I[\X, \Z; q_i]
\end{equation}
where $\E{m}$ is the expected number of messages in the server and $I[\X, \Z; q_i]$ is the information for a single message.
Using~\eqref{eq:coalitioninfo} one can find $I[\X, \Z; q_i]$, noting that $\alpha_C = \overline{p}_d^{|C|} \lambda$ and $\beta = \lambda \overline{p}_d \log \overline{p}_d$ since $p_d$ does not depend on $x$.

It only remains to model the expected number of new measurements available in the server.
Assuming that there is an average rate of return, $\rho \approx 1/T_C$, then a geometric distribution models the discrete waiting time between events.
The number of messages in the server will be equal to $\tau_i - k$, where $\tau_i$ is the number of time steps since the robot under consideration last communicated with the server (\ie the length of the local message history) and $k$ is the number of time steps for another robot.
Finally, assuming robots move independently, since there are $N-1$ other robots we have:
\begin{equation}
\E{m} = (N-1) \sum_{k = 0}^{\tau_i} (\tau_i-k) (1-\rho)^k \rho.
\end{equation}

%%%%%%%%%%%%%%%%%%%%%%%%%%%%%%
\subsection{Computation Complexity}
\label{sec:complexity}
%%%%%%%%%%%%%%%%%%%%%%%%%%%%%%
While aspects of the computational complexity of the algorithm have been hinted at, we formally address the issue here.
As written, the complexity of the mutual information computations in \eqref{eq:controllaw} for a single robot in coalition $C$ is $\mathcal{O}(P~2^{|C|}~\prod_{i \in C}|\G_i|)$, where $P$ is the number of particles used to represent the PHD, the factor of $2^{|C|}$ comes from the possible binary measurement vectors, and the remaining term is for the possible coalition configurations.
Note that this is exponential in the size of the coalition both through measurements and actions, which motivates the use of both the binary measurement model and the control graph $\G$ to keep the computations tractable.

%%%%%%%%%%%%%%%%%%%%%%%%%%%%%%%%%%%%%%%%%%%%%%%%%%%%%%%%%%%
\section{Results}
\label{sec:results}
\begin{figure}[th]
\centering
\includegraphics[width=0.75\columnwidth]{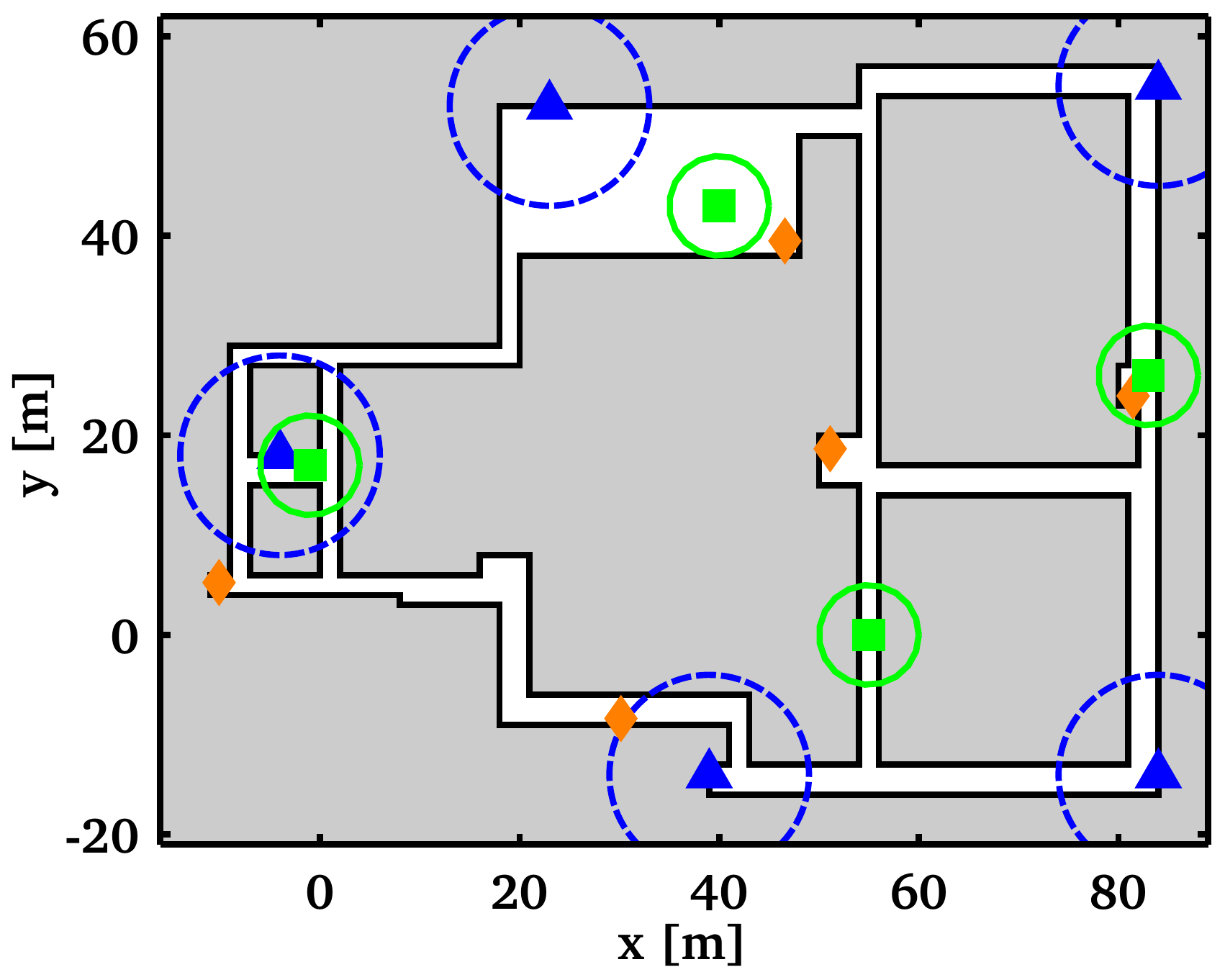} \vspace*{-2mm}
\caption{Example environment with four robots (green squares) shown with their sensor footprints (green circles).
There are five targets (orange diamonds) and five access points (blue triangles), which have limited communication range (dashed blue circles), within the environment.}
\label{fig:exampleworld}
\vspace*{-3mm}
\end{figure}

%%%%%%%%%%%%%%%%%%%%%%%%%%%%%%%%%%%%%%%%%%%%%%%%%%%%%%%%%%%
The example scenario considered here involves a team of four mobile robots searching for targets within a large indoor office environment, as shown in Fig.~\ref{fig:exampleworld}.
Robots are equipped with omnidirectional sensors with circular footprints (of radius $r_d$) and probability of detection given by
\begin{equation}
p_d(x; q_i) = \begin{cases}
       p_{d, 0} e^{-|x-q_i|^2/\sigma_d^2} & \text{if $|x-q_i| \leq r_d$} \\
       0 & \text{if $|x-q_i| > r_d$}
     \end{cases}
\end{equation}
where $p_{d,0} = 0.8$, $\sigma_d = 2$~m, and $r_d = 5$~m.
The measurement model is given by
\begin{equation}
g(z \mid x) = x + \eta
\end{equation}
where $\eta \sim \mathcal{N}(0, \sigma_g^2)$ is Gaussian white noise with $\sigma_g = 1$~m.
The expected number of clutter points in the footprint is $\mu = 0.3$ and $\kappa(z) = \mu / |F|$ is uniform over the footprint.

Nodes in the control graph $\G$ form a uniform grid with 1~m spacing.
We assume minimal kinematic restrictions and say that robots may move up to 2~m in any direction in a single time step, so the graph is 12-connected.
The PHD is represented by a set of uniformly spaced particles in a 1~m grid on the robots and a 0.2~m grid on the server and $\lambda$ is initially set to 20.

%%%%%%%%%%%%%%%%%%%%%%%%%%%%%%
%%%%%%%%%%%%%%%%%%%%%%%%%%%%%%

There are five access points within the environment and we use a simple disk model for communication, with access points and robots having a range of 10~m.
The check-in time, $T_C$, is set to 40 time steps, well above the minimum number of motions, 23, required to reach any point in the environment from its nearest access point.

Using this setup we simulate the system for 1000 time steps, with the team often finding all the targets and localizing them to within 0.5~m accuracy.
To extract the final target estimate from the PHD, we use a simple thresholding and clustering scheme.
First, any point with PHD smaller than some $w_{\rm min} \ll 1$ (we use 0.02) is ignored.
From the remaining points we find clusters with total weight above 0.5, where nodes are connected if they are within an 8-connected neighborhood of one another.
Finally, the expected locations are the weighted mean of the particles in each cluster.
From a typical trial, the errors in localizing true targets were $\{0.09, 0.21, 0.29, 0.33, 0.88\}$~m, all less than both the grid size and the standard deviation of the sensor noise.
In the same trial there was one false positive target, due to clutter detections while a robot was passing through the hallway in Explore mode with no robot returning to investigate before the simulation ended.
Fig.~\ref{fig:modessinglerun} shows the time evolution of the control modes for each robot.

%%%%%%%%%%%%%%%%%%%%%%%%%%%%%%
\subsection{Key System Parameters}
%%%%%%%%%%%%%%%%%%%%%%%%%%%%%%
There are several key parameters that influence the behavior of the robot team.
Namely, the number of robots $N$, the characteristic length of the sensors $R_S$, the maximum robot velocity $V$, the number of access points $A$, the communication range $R_C$, the check-in time $T_C$, and the characteristic length of the environment $L$.

\begin{figure}[th]
\centering
\includegraphics[width=0.9\columnwidth]{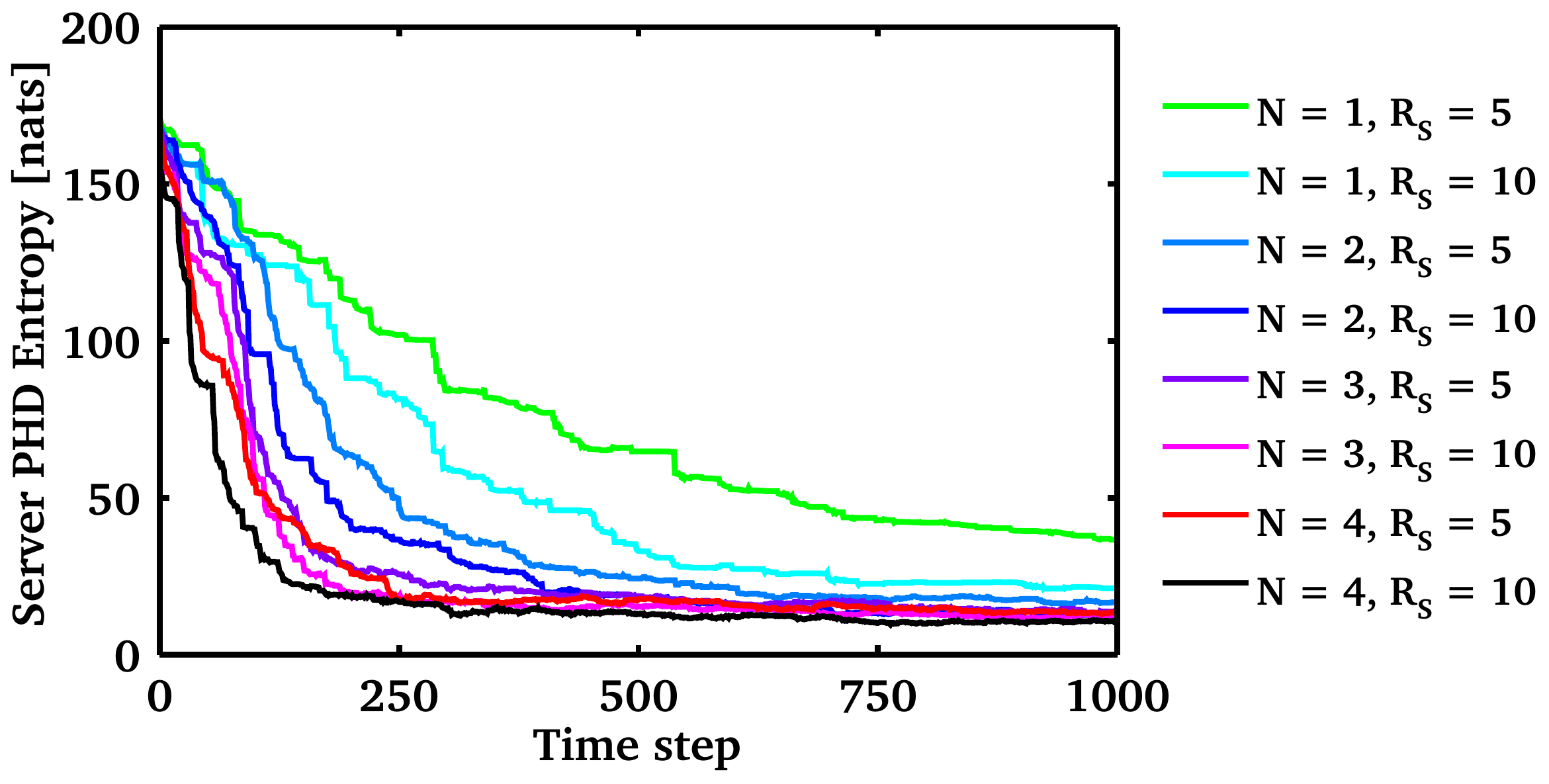} \vspace*{-3mm}
\caption{Time evolution of the entropy of the target RFS for a variety of team sizes and footprint radii.}
\label{fig:entropy}
%\vspace*{-4mm}
\end{figure}

The fraction of information retrieved per time step decreases with the size of the environment, $L$, but it can be explored more quickly by using more robots, $N$, or increasing the visible area per robot, $R_S$.
To investigate the effects of $N$ and $R_S/L$ on the rate of information retrieval, we conducted a series of simulations using between 1 and 4 robots and two footprint radii, 5 and 10~m, with 10 trials for each set of parameters.
The resulting time-evolution of the average entropy (a measure of uncertainty) of the server PHD is shown in Fig.~\ref{fig:entropy}.
As is expected, a higher number of robots and a larger sensing radius both lead to a higher rate of information gathering, as evidenced by the lower entropy.
The entropy of a Poisson random finite is given by
\begin{equation}
\label{eq:targetentropy}
H[X] = \lambda \big(1 - \log \lambda - H[d(x)] \big)
\end{equation}
where $d(x) = \lambda^{-1} D(x)$ is the normalized PHD.
See Appendix~\ref{app:entropy} for the derivation.

%%%%%%%%%%%%%%%%%%%%%%%%%%%%%%
%%%%%%%%%%%%%%%%%%%%%%%%%%%%%%

\begin{figure}[t]
\centering
\subfloat[Single run]{
	\includegraphics[width=0.48\columnwidth]{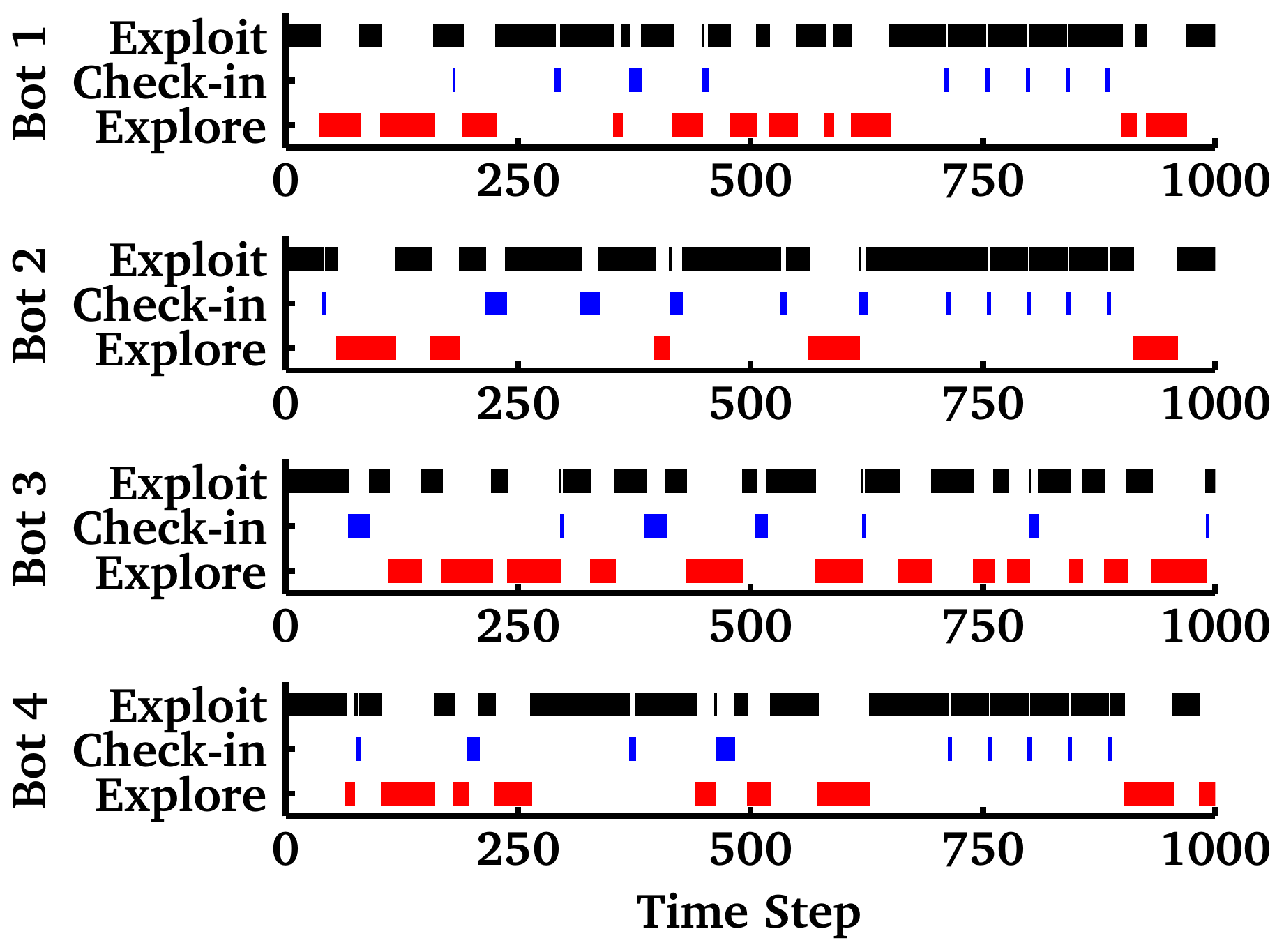}
	\label{fig:modessinglerun}
}
\subfloat[As a function of $T_C$]{
	\includegraphics[width=0.46\columnwidth]{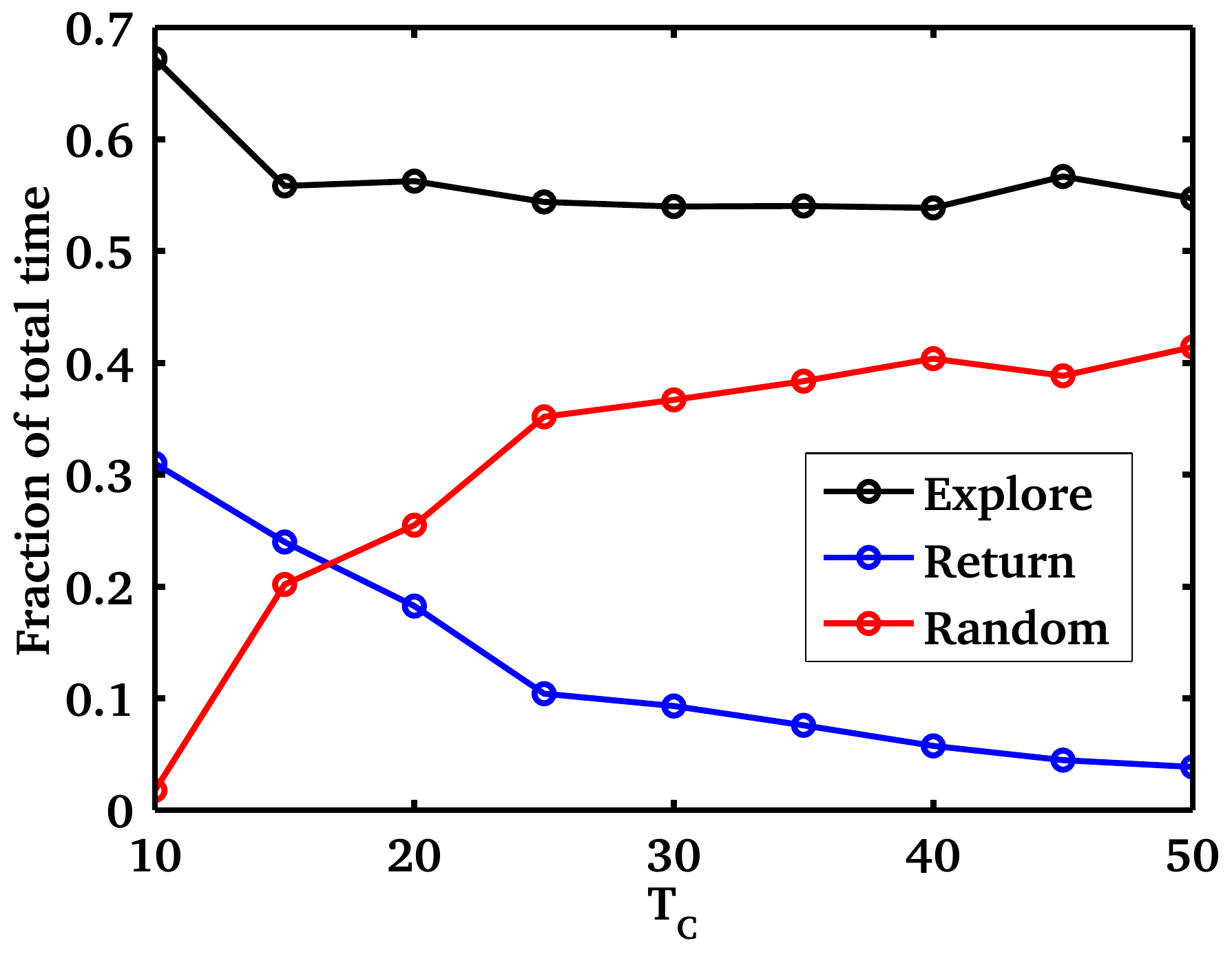}
	\label{fig:modesaverage}
}
\caption{Time spent in each control mode, Exploit (black), Check-in (blue), and Explore (red).
(a) The time evolution of the mode switching for each individual robot over an example run.
(b) The fraction of the total time spent in each mode as a function of $T_C$.}
\label{fig:controlmodes}
\vspace*{-4mm}
\end{figure}

As the environment grows in size, the time between uploads to the server, $T_C$, must increase so that robots are able to reach more distant locations.
Conversely, robots are able to reach an access point more quickly as the access point density $A/L^2$, communication range $R_C$, and robot speed $V$ all increase.
To investigate the effects of this exploration time on the system behavior, we ran a series of simulations varying $T_C$ from 10 to 50 time steps by increments of 5, with 10 trials for each rate.
The average fraction of the total simulation time spent in each control mode is shown in Fig.~\ref{fig:modesaverage}.
For obvious reasons, it is desirable for the fraction of time spent in the Exploit mode to be as high as possible because this means the robots do not spend large amounts of time driving to access points or getting stuck.
Not surprisingly, as the check-in rate decreases, the fraction of the total time spent in Check-in mode also decreases.
On the other hand, as the ratio of $T_C$ to $T_S$ increases the robot gets stuck more often so it spends more time in Explore mode.
The surprising thing is that these two effects appear to cancel one another out, with the total fraction of the times spent exploring at around 0.55 for every value of $T_C$ except $T_C = 10$.

%%%%%%%%%%%%%%%%%%%%%%%%%%%%%%
\subsection{Cooperation}
%%%%%%%%%%%%%%%%%%%%%%%%%%%%%%
One obvious question to ask is how much benefit leader election within a coalition provides, as opposed to allowing each robot to redundantly plan the coalition action based on its own PHD.
In other words, does having different PHDs among the coalition members hurt the performance of the team.
To explore this issue we ran another series of simulations where robots did not run the leader election policy.
Instead each robot redundantly planned the action of the coalition, effectively acting as the leader but not sharing these plans with other robots.

\begin{figure}[t]
\centering
\subfloat[With leader election]{
	\includegraphics[width=0.47\columnwidth]{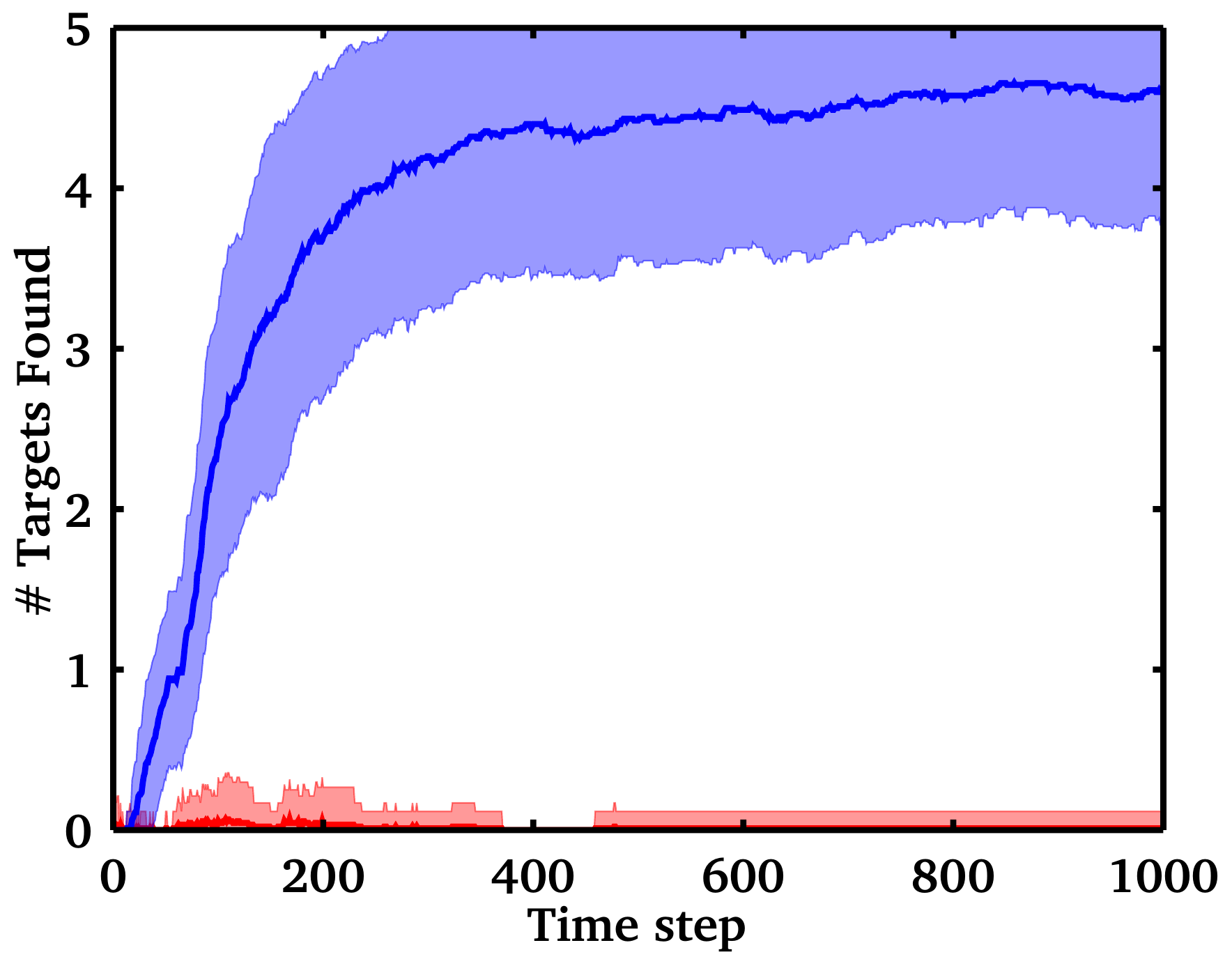}
}
\subfloat[Without leader election]{
	\includegraphics[width=0.47\columnwidth]{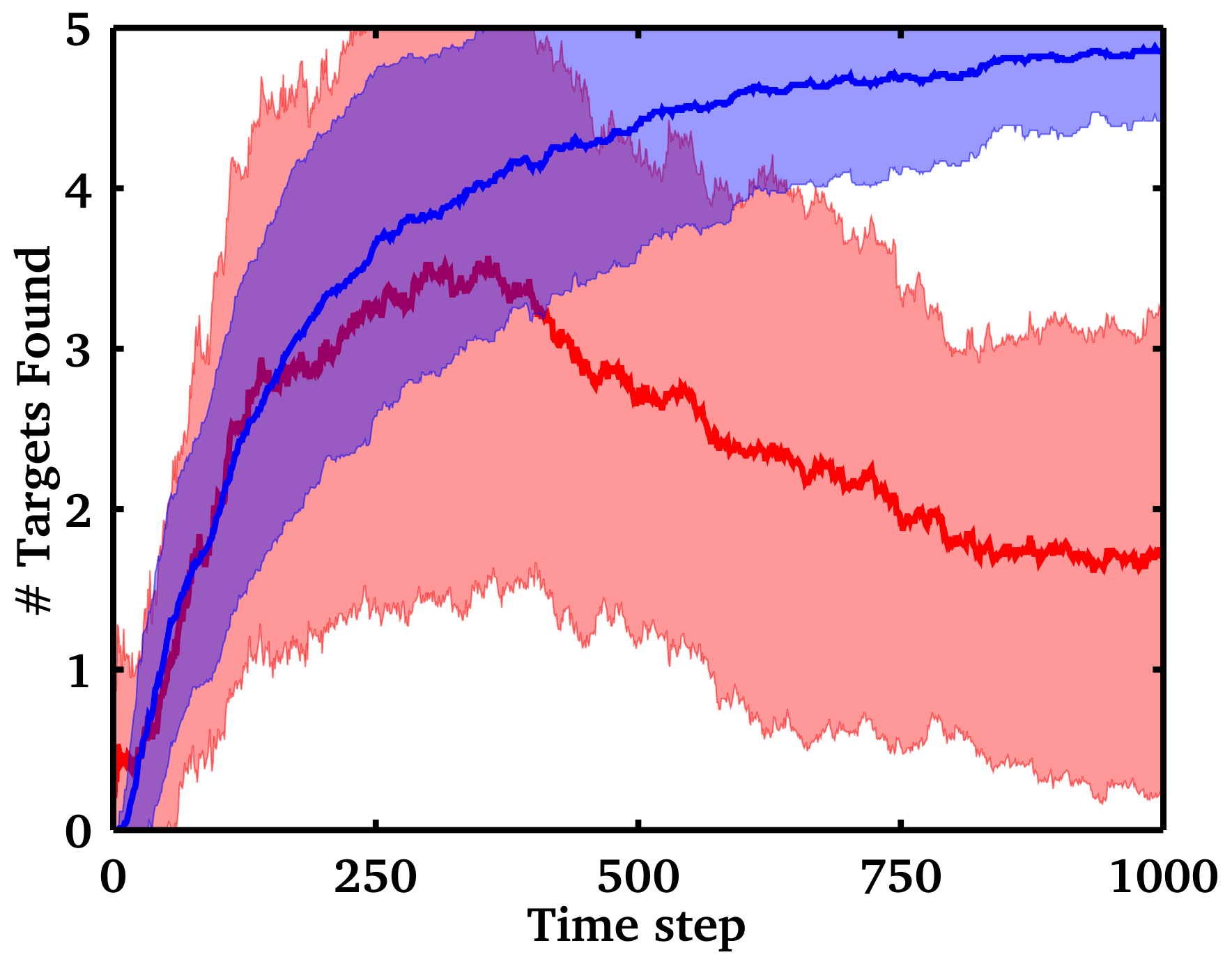}
}
\caption{Plots showing the time evolution of the number of true targets (blue) and false targets (red).
The mean over 90 separate trials is shown by the solid line and the shaded regions correspond show one standard deviation.
}
\label{fig:targetsfound}
\vspace*{-4mm}
\end{figure}

The major difference between the two modes was the rate at which false positive targets arise, as shown in Fig.~\ref{fig:targetsfound}.
While the mean value and standard deviation of true targets are quite similar, the team without the leader election policy has a significantly higher rate of false positive targets.
This indicates that one of the primary benefits of leader election is for error mitigation: robots tend to get in each others way or not move in complementary directions when they plan based on different PHDs.

Finally, we return to the issue of computational complexity from Sec.~\ref{sec:complexity}.
In our simulations, run in Matlab on a laptop with a 2.27~GHz Intel Core i3 with 4~GB of RAM, mutual information for coalition of a single robot took an average of 0.014~s to compute, of two robots an average of 0.484~s, and of three robots an average of 11.829~s.
Real-time implementation of this system was not the subject of this work, with these numbers meant to indicate the feasibility, for example using C\verb!++! could likely reduce the computation time by an order of magnitude and using a GPU could reduce it significantly more, as mutual information is highly parallelizable.
Implementation of the system in hardware will be the study of future work.

%%%%%%%%%%%%%%%%%%%%%%%%%%%%%%
%%%%%%%%%%%%%%%%%%%%%%%%%%%%%%

%%%%%%%%%%%%%%%%%%%%%%%%%%%%%%
%%%%%%%%%%%%%%%%%%%%%%%%%%%%%%

%%%%%%%%%%%%%%%%%%%%%%%%%%%%%%%%%%%%%%%%%%%%%%%%%%%%%%%%%%%
\section{Conclusion}
\label{sec:conclusion}
%%%%%%%%%%%%%%%%%%%%%%%%%%%%%%%%%%%%%%%%%%%%%%%%%%%%%%%%%%%
In this paper we propose a cooperative exploration strategy for multi-target localization with noisy sensors.
Estimation is done using the PHD filter, allowing the inclusion of false positive and false negative detections, high sensor noise, and unknown data association in a principled way.
We also describe a network architecture wherein robots exchange information on a peer-to-peer basis as well as communicate with a central server.
The server allows information to be shared with robots potentially exploring disjoint regions of the environment, synchronizes the belief across the team, and potentially allows the robots access to cloud services.
Our proposed control law is based on maximizing mutual information between the target set and sensor measurements, both from on-board sensors of robots in the coalition and the expected measurements uploaded to the server by other robots, combining the goals of information collection and dissemination into a single objective function.
To the authors' best knowledge, no such expressions for mutual information based upon the PHD filter have previously appeared in the literature.
Furthermore, robots in the same region of the environment form coalitions and plan joint actions in order to cooperatively localize targets.
Finally, we present extensive simulation results in a complex, indoor environment, studying the effects of varying multiple system parameters on the performance of the team and demonstrating the potential for real-time implementation.

%%%%%%%%%%%%%%%%%%%%%%%%%%%%%%%%%%%%%%%%%%%%%%%%%%%%%%%%%%%
\appendices
\input{derivationMI}

\end{document}

%% file: derivationMI.tex
%%%%%%%%%%%%%%%%%%%%%%%%%%%%%%
\section{Mutual Information of a Signle Binary Measurement and a Poisson RFS}
\label{app:MIsingle}
%%%%%%%%%%%%%%%%%%%%%%%%%%%%%%
From the definition of mutual information, in~\eqref{eq:coalitioninfo} to~\eqref{eq:conditionalentropy}, there are two main terms of interest the entropy and conditional entropy.
Here we consider the single-robot case and extend this to multiple robots in Appendix~\ref{app:MImultiple}.
%Let $p(z_0) \triangleq p(z = 0)$.
Let us first look at the entropy
\[
H[\Z] = - \sum_{z = \{0,1\}} p_b(z) \log p_b(z). \tag{\ref{eq:entropy}}
\]
This only requires us to calculate $p_b(z=0)$ since $p_b(z=1) = \pbar_b(z=0)$.
From the definition of the measurement model and the target distribution, we have
\begin{align}
p_b(z=0) = & \, \int p_b(z = 0 \mid X) p(X) \, \delta X \label{eq:singlenodetect} \\
	= & \, \sum_{k = 0}^\infty \frac{1}{k!} \int p_K(0) e^{-\lambda} \notag \\
		& \quad \times \prod_{j=1}^k \pbar_d(x_j) D(x_j) \, dx_1 \ldots dx_k \notag \\
	= & \, p_K(0) e^{-\lambda} \sum_{k = 0}^\infty \frac{1}{k!} \prod_{j=1}^k \underbrace{\int \pbar_d(x_j) D(x_j) \, dx_j}_\alpha \notag \\
	= & \, e^{-\mu} e^{-\lambda} e^\alpha \sum_{k = 0}^\infty \frac{1}{k!} e^{-\alpha} \alpha^k \notag \\
	= & \, e^{-(\lambda - \alpha + \mu)} \notag
\end{align}
since the sum is simply the total probability mass of a Poisson random variable with parameter $\alpha$, which is identically 1.
Note that this is guaranteed to be a probability since $\alpha \leq \lambda$ so the exponent is non-negative.
Then using the additive complement, $p_b(1) = 1 - e^{-(\lambda - \alpha + \mu)}$.

Note the definitions of $\lambda$ and $\alpha$ are very similar, with $\alpha$ being the same as $\lambda$ weighted by the probability of no detection at each location.
This leads to the nice interpretation of $\lambda - \alpha$ as the expected number of detected targets while $\mu$ is the expected number of false positive detections.

Next we look at the conditional entropy computations.
\[
H[\Z \mid \X] = -\sum_{z \in \{0,1\}} \int p_b(z,X) \log p_b(z \mid X) \, \delta X.
	\tag{\ref{eq:singlecondent}}
\]
Beginning with the $z=0$ term of the conditional entropy,
\begin{align}
		& \int p_b(z=0 \mid X) p(X) \log p_b(z=0 \mid X) \, \delta X \notag \\
			&\hspace*{3em} = \sum_{k = 0}^\infty \frac{1}{k!} \int p_K(0) e^{-\lambda} \prod_{j=1}^k \pbar_d(x_j) D(x_j) \notag \\
			&\hspace*{3em} \qquad \times \log p_K(0) \prod_{i=1}^k \pbar_d(x_i) \, dx_1 \ldots dx_k \notag \\
		&\hspace*{3em} = e^{-\mu} e^{-\lambda} \sum_{k = 0}^\infty \frac{1}{k!} \int \prod_{j=1}^k \pbar_d(x_j) D(x_j) \notag \\
			&\hspace*{3em} \qquad \times \bigg[ \log e^{-\mu} + \sum_{i=1}^k \log \pbar_d(x_i) \bigg] \, dx_1 \ldots dx_k \notag \\
		&\hspace*{3em} = e^{-(\lambda+\mu)} \sum_{k = 0}^\infty \frac{1}{k!} \bigg[ -\mu \alpha^k \notag \\
			&\hspace*{3em} \qquad + k \alpha^{k-1} \underbrace{\int \pbar_d(x) D(x) \log \pbar_d(x) \, dx}_{-\beta} \bigg] \notag \\
		&\hspace*{3em} = - e^{-(\lambda - \alpha + \mu)} \big(\beta + \mu \big) \sum_{k = 0}^\infty e^{-\alpha} \frac{1}{k!} \alpha^k \notag \\
		&\hspace*{3em} = - e^{-(\lambda - \alpha + \mu)} \big( \beta + \mu \big) \label{eq:condent0}
	\end{align}
where we again note that we have the total probability mass of a Poisson random variable.

The final term is the only one that does not have a nice, closed-form solution like the previous ones, due to the presence of a sum inside the log term:
	\begin{multline*}
		\int p_b(z=1 \mid X) p(X) \log p_b(z=1 \mid X) \, \delta X \\
			= \int (1 - p_b(z=0 \mid X)) p(X) \log (1- p_b(z=0 \mid X)) \, \delta X.
	\end{multline*}
To get around this, we take the Taylor series of $\log(1-p_0)$ about $p_0 = 0$, where $p_0 = p_b(z=0 \mid X)$ for compactness, so:
\[
\log(1-p_0) \approx -p_0 - \frac{1}{2} p_0^2 - \frac{1}{3} p_0^3 + \ldots
\]
Substituting this into the integral, we have
\begin{align}
&\int p_b(z=1 \mid X) p(X) \log p_b(z=1 \mid X) \, \delta X \notag \\
	& \quad \approx \int (1-p_0) (-p_0 - \frac{1}{2} p_0^2 - \frac{1}{3} p_0^3 + \ldots) p(X) \, \delta X \notag \\
	& \quad = \sum_{\ell = 1}^\infty c_\ell \int p_0^\ell p(X) \, \delta X \notag
\end{align}
where $c_\ell$ are the coefficients, which are $c_\ell = 1/(\ell(\ell-1))$ for $\ell > 1$ and $c_1 = -1$.
We can now plug in the definitions of $p_0$ and $p(X)$ to get
\begin{align}
& \int p_b(z=1 \mid X) p(X) \log p_b(z=1 \mid X) \, \delta X \notag \\
	& \quad \approx \sum_{\ell = 1}^\infty c_\ell \sum_{k=0}^\infty \frac{1}{k!} \int p_K(0)^\ell e^{-\lambda} \notag \\
	& \qquad \times \prod_{j=1}^k \pbar_d(x_j)^\ell D(x_j) \, dx_1 \ldots dx_k \notag \\
	& \quad \approx \sum_{\ell = 1}^\infty c_\ell e^{-\ell \mu} e^{-(\lambda - \alpha(1^\ell))} \sum_{k=0}^\infty \frac{1}{k!} e^{-\alpha(1^\ell)} (\alpha(1^\ell))^k \notag \\
	& \quad \approx \sum_{\ell = 1}^\infty c_\ell e^{-(\lambda - \alpha(1^\ell) + \ell \mu)} \label{eq:condent1}
\end{align}
where $1^\ell = \{1, \ldots, 1\}$ is a set containing $\ell$ copies of 1 and $\alpha(1^\ell) = \int \pbar_d(x)^\ell D(x) \, dx$.
Then we can see that \eqref{eq:conditionalentropy} is the sum of \eqref{eq:condent0} and \eqref{eq:condent1}.

%%%%%%%%%%%%%%%%%%%%%%%%%%%%%%
\section{Mutual Information of Multiple Binary Measurements and a Poisson RFS}
\label{app:MImultiple}
%%%%%%%%%%%%%%%%%%%%%%%%%%%%%%
The approach from Appendix~\ref{app:MIsingle} can be easily extended to work with multiple robots, assuming conditional independence of sensor measurements given the environment.
This conditional independence  results in the conditional entropy of the team simply being the sum of the conditional entropies of each robot.
This allows us to simply write \eqref{eq:coalitioncondent} from \eqref{eq:singlecondent}.

The entropy terms do not decouple as nicely, and the computational complexity will be exponential in the number of robots.
Here we must compute $p_b(Z)$ for each random vector $Z$ of sensor measurements.
Let the robots under consideration be in coalition $C$ and let $C^0 = \{j \in C \mid z_j = 0\}$ be all the robots without a detection and $C^1 = \{j \in C \mid z_j = 1\} = C \setminus C^0$ be all the robots with a detection.
Then, letting $p_i(0) = p_b(z_i = 0 \mid X)$, we get
	\begin{align*}
		p_b(Z) = & \int \prod_{i\in C} p_i(z) \, p(X) \, \delta X \\
			= & \int \prod_{i \in C^0} p_i(0) 
			 	\prod_{j \in C^1} (1-p_j(0)) \, p(X) \, \delta X \\
			= & \int \prod_{i \in C^0} p_i(0) \sum_{C' \subseteq C^1} (-1)^{|C'|} \prod_{j \in C'} p_j(0) \, p(X) \, \delta X
	\end{align*}
where the last equality comes from expanding the product over $C^1$.
Plugging in the definitions of the measurement models and target distributions, we have an integral of the same form as~\eqref{eq:singlenodetect}.
Through identical arguments to those in Appendix~\ref{app:MIsingle}, we get:
\[
p_b(Z) = \sum_{C' \subseteq C^1} (-1)^{|C'|} e^{-(\lambda + \mu |C^0 \cup C'| - \alpha(C^0 \cup C'))}. \tag{\ref{eq:measurementlikelihood}}
\]

%%%%%%%%%%%%%%%%%%%%%%%%%%%%%%
\section{Entropy of Poisson RFS}
\label{app:entropy}
%%%%%%%%%%%%%%%%%%%%%%%%%%%%%%
We wish to find an expression for the entropy of the target set, given the likelihood function:
\[
p(X) = e^{-\lambda} \prod_{x \in X} D(x). \tag{\ref{eq:poissonrfs}}
\]
Plugging this into the standard formula, we get:
	\begin{align}
		H[\mathcal{X}] &= - \int p(X) \log p(X) \, \delta X \notag \\
			&= - \int e^{-\lambda} \prod_{x \in X} D(x) \log \bigg[ e^{-\lambda} \prod_{x \in X} D(x) \bigg] \, \delta X \notag \\
			&= -e^{-\lambda} \sum_{k=0}^\infty \frac{1}{k!} \int \prod_{i=1}^k D(x_i) \notag \\
				& \quad \times \bigg[ -\lambda + \sum_{j=1}^k \log D(x_j) \bigg] \, dx_1 \ldots dx_k \notag \\
			&= -e^{-\lambda} \sum_{k=0}^\infty \frac{1}{k!} \bigg[ -\lambda \bigg( \int D(x) \, dx \bigg)^k \notag \\
				& \quad + k \bigg( \int D(x) \, dx \bigg)^{k-1} \bigg( \int D(x) \log D(x) \, dx \bigg) \bigg] \notag \\
			&= \bigg( \lambda - \int D(x) \log D(x) \, dx \bigg) \sum_{k=0}^\infty \frac{1}{k!} \lambda^k e^{-\lambda} \notag \\
			&= \lambda - \int D(x) \log D(x) \, dx. \notag
	\end{align}
This may also be written in terms of the normalized density $d(x) = \lambda^{-1} D(x)$,
	\begin{align}
		H[\mathcal{X}] &= \lambda - \lambda \int d(x) [\log \lambda + \log d(x)] \, dx \notag \\
			&= \lambda - \lambda \log \lambda - \lambda \int d(x) \log d(x) \, dx \notag \\
			&= \lambda (1 - \log \lambda + H[d(x)]). \tag{\ref{eq:targetentropy}}
	\end{align}